\documentclass[10pt,twocolumn,letterpaper]{article}

\usepackage{iccv}
\usepackage{times}
\usepackage{epsfig}
\usepackage{graphicx}
\usepackage{amsmath}
\usepackage{amssymb}
\usepackage{subcaption}
\usepackage{multirow}
\usepackage{stfloats}

\usepackage[pagebackref=true,breaklinks=true,letterpaper=true,colorlinks,bookmarks=false]{hyperref}

\iccvfinalcopy 

\def\iccvPaperID{2005} 

\ificcvfinal\fi
\begin{document}

\title{Scale-Aware Trident Networks for Object Detection}

\author{
Yanghao Li*\quad 
Yuntao Chen$^{1,3}$*\quad 
Naiyan Wang$^{2}$ \quad 
Zhaoxiang Zhang$^{1,3,4}$ \\
$^{1}$ University of Chinese Academy of Sciences \qquad
$^{2}$ TuSimple\\
$^{3}$ Center for Research on Intelligent Perception and Computing, CASIA\\
$^{4}$Center for Excellence in Brain Science and Intelligence Technology, CAS\\
{\tt\small
\{lyttonhao, winsty\}@gmail.com \{chenyuntao2016, zhaoxiang.zhang\}@ia.ac.cn}
}

\maketitle
{\let\thefootnote\relax\footnote{* Equal Contribution}}
\graphicspath{{figures/}}

\begin{abstract}

Scale variation is one of the key challenges in object detection. 
In this work, we first present a controlled experiment to investigate the effect of receptive fields for scale variation in object detection.
Based on the findings from the exploration experiments, we propose a novel Trident Network (TridentNet) aiming to generate scale-specific feature maps with a uniform representational power. 
We construct a parallel multi-branch architecture in which each branch shares the same transformation parameters but with different receptive fields.
Then, we adopt a scale-aware training scheme to specialize each branch by sampling object instances of proper scales for training. 
As a bonus, a fast approximation version of TridentNet could achieve significant improvements without any additional parameters and computational cost compared with the vanilla detector. 
On the COCO dataset, our TridentNet with ResNet-101 backbone achieves state-of-the-art single-model results of \textbf{48.4 mAP}.
Codes are available at \url{https://git.io/fj5vR}.

\end{abstract}

\begin{section}{Introduction}

In recent years, deep convolutional neural networks (CNNs)~\cite{rcnn,faster-rcnn,ssd}  have achieved great success in object detection.
Typically, these CNN-based methods can be roughly divided into two types: one-stage methods such as YOLO~\cite{yolo} or SSD~\cite{ssd} which directly utilizes feed-forward CNN to predict the bounding boxes of interest, while two-stage methods such as Faster R-CNN~\cite{faster-rcnn} or R-FCN~\cite{rfcn} first generate proposals, and then exploit the extracted region features from CNN for further refinement.
However, a central issue for both methods is the handling of scale variation. 
The scales of object instances could vary in a wide range, which impedes the detectors, especially those very small or very large ones. 


To remedy the large scale variation, an intuitive way is to leverage multi-scale image pyramids~\cite{adelson1984pyramid}, which is popular in both hand-crafted feature based methods~\cite{dalal2005histograms,lowe2004distinctive} and current deep CNN based methods (Figure~\ref{fig:architectures}(a)). 
Strong evidence~\cite{huang2017speed,liu2018path} shows that deep detectors~\cite{faster-rcnn,rfcn} could benefit from multi-scale training and testing. 
To avoid training objects with extreme scales (small/large objects in smaller/larger scales), SNIP~\cite{snip,sniper} proposes a scale normalization method that selectively trains the objects of appropriate sizes in each image scale. 
Nevertheless, the increase of inference time makes the image pyramid methods less favorable for practical applications.

Another line of efforts aims to employ in-network feature pyramids to approximate image pyramids with less computation cost. 
The idea is first demonstrated in~\cite{dollar2014fast}, where a fast feature pyramid is constructed for object detection by interpolating some feature channels from nearby scale levels. 
In the deep learning era, the approximation is even easier. 
SSD~\cite{ssd} utilizes multi-scale feature maps from different layers and detects objects of different scales at each feature layer. 
To compensate for the absence of semantics in low-level features, FPN~\cite{fpn} (Figure~\ref{fig:architectures}(b)) further augments a top-down pathway and lateral connections to incorporate strong semantic information in high-level features. 
However, region features of objects with different scales are extracted from different levels of FPN backbone, which in turn are generated with different sets of parameters.
This makes feature pyramids an unsatisfactory alternative for image pyramids.

\begin{figure*}[t]
\begin{subfigure}{0.33\linewidth}
\centering
\includegraphics[width=1.0\linewidth]{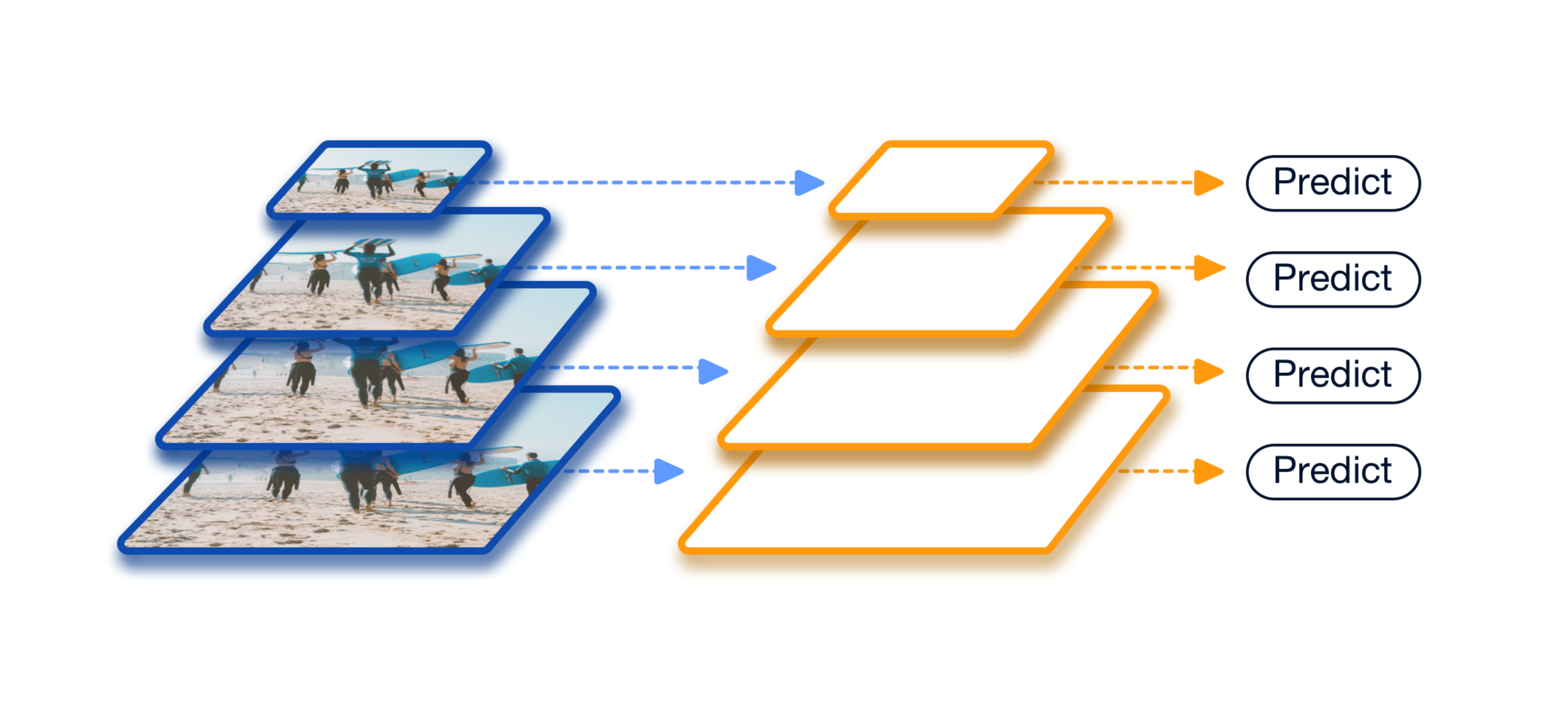}
\caption{Image Pyramid}\label{fig:image_pyramid}
\end{subfigure}
\begin{subfigure}{0.33\linewidth}
\centering
\includegraphics[width=1.0\linewidth]{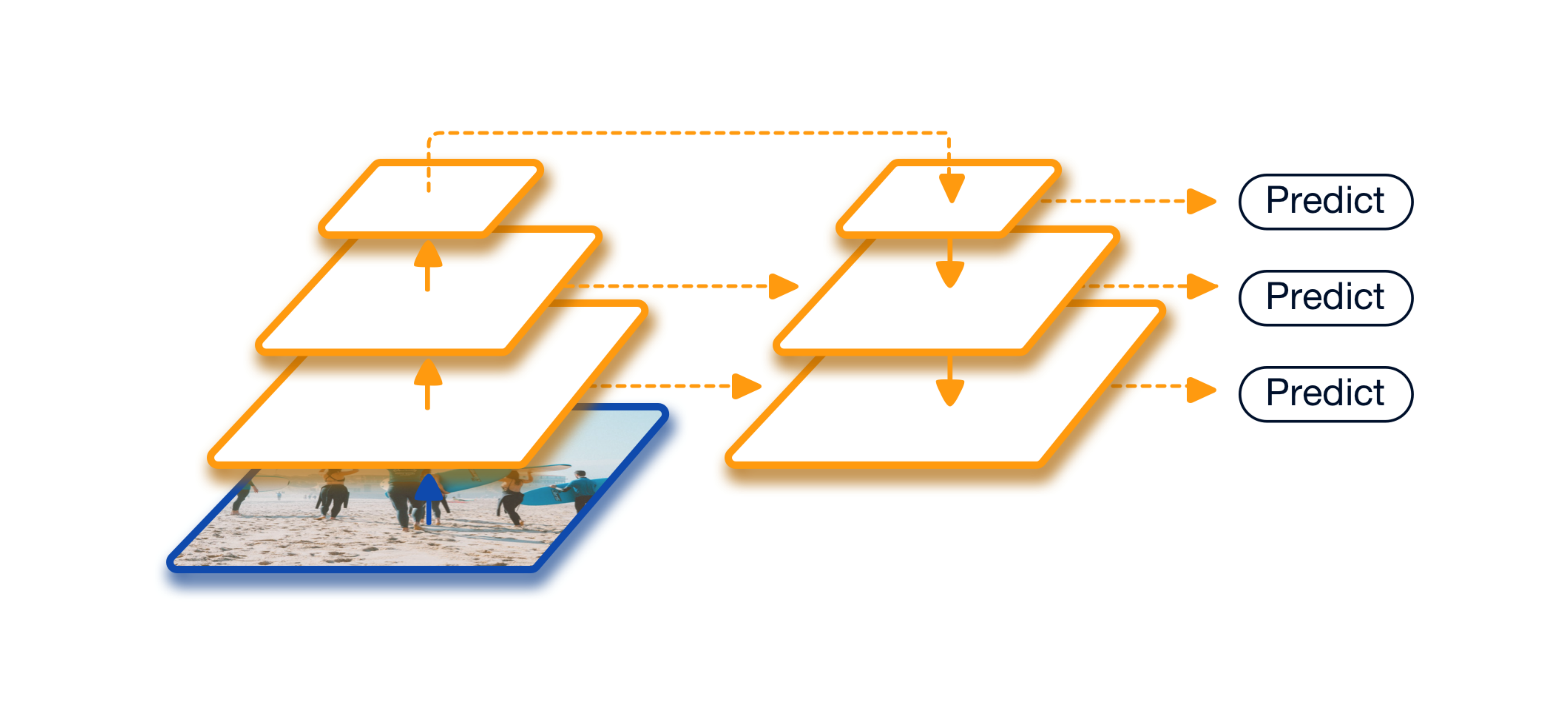}
\caption{Feature Pyramid}\label{fig:feature_pyramid}
\end{subfigure}%
\begin{subfigure}{0.33\linewidth}
\centering
\includegraphics[width=1.0\linewidth]{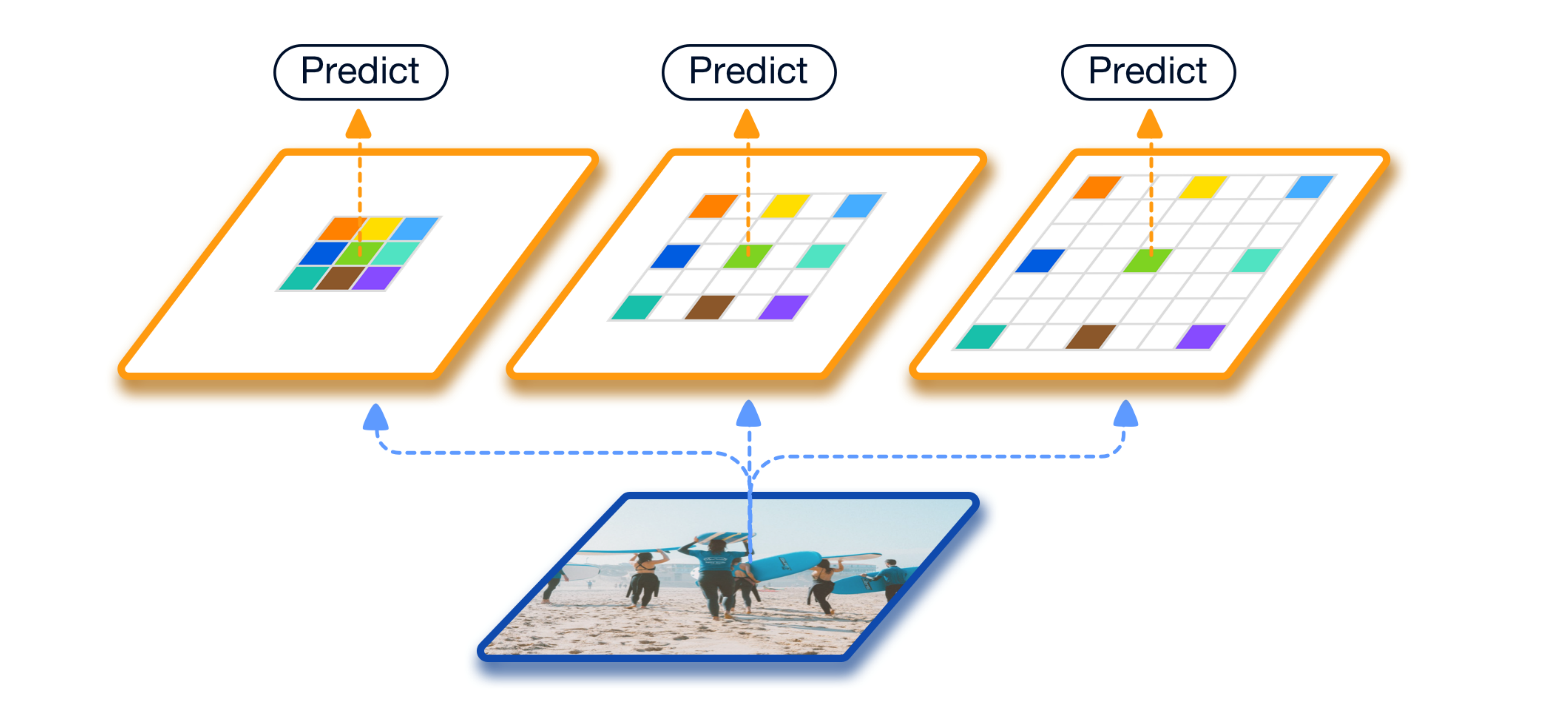}
\caption{Trident Network}\label{fig:trident}
\end{subfigure}%
\vspace{-1mm}
\caption{(a) Using multiple images of several scales as input, the image pyramid methods perform feature extraction and object detection independently for each scale. (b) The feature pyramid methods utilize the features from different layers of CNNs for different scales, which is computational friendly. This figure takes FPN~\cite{fpn} as an example. (c) Our proposed Trident Network generates scale-aware feature maps efficiently by trident blocks with different receptive fields.}\label{fig:architectures}
\end{figure*}

Both the image pyramid and the feature pyramid methods share the same motivation that models should have different receptive fields for objects of different scales. 
Despite the inefficiency, the image pyramid fully utilizes the representational power of the model to transform objects of all scales equally. 
In contrast, the feature pyramid generates multi-level features thus sacrificing the feature consistency across different scales. This leads to a decrease in effective training data and a higher risk of overfitting for each scale.
The goal of this work is to get the best of two worlds by creating features with a uniform representational power for all scales efficiently.

In this paper, instead of feeding in multi-scale inputs like the image pyramid, we propose a novel network structure to adapt the network for different scales. 
In particular, we create multiple scale-specific feature maps with the proposed trident blocks as shown in Figure~\ref{fig:architectures}(c). With the help of dilated convolutions~\cite{yu2016multi}, different branches of trident blocks have the same network structure and share the same parameters yet have different receptive fields. Furthermore, to avoid training objects with extreme scales, we leverage a scale-aware training scheme to make each branch specific to a given scale range matching its receptive field. Finally, thanks to weight sharing through the whole multi-branch network, we could approximate the full TridentNet with only one major branch during inference. This approximation only brings marginal performance degradation. As a result, it could achieve significant improvement over the single-scale baseline without any compromise of inference speed. This property makes TridentNet more desirable over other methods for practical uses. 

To summarize, our contributions are listed as follows:
\begin{itemize}
\item We present our investigation results about the effect of the receptive field in scale variation. To our best knowledge, we are the first to design controlled experiments to explore the receptive field on the object detection task.
\item We propose a novel Trident Network to deal with scale variation problem for object detection. Through multi-branch structure and scale-aware training, TridentNet could generate scale-specific feature maps with a uniform representational power. 
\item We propose a fast approximation, TridentNet Fast, with only one major branch via our weight-sharing trident-block design, thus introducing no additional parameters and computational cost during inference.
\item We validate the effectiveness of our approach on the standard COCO benchmark with thorough ablation studies. Compared with the state-of-the-art methods, our proposed method achieves \textbf{an mAP of 48.4 using a single model with ResNet-101 backbone}. 
\end{itemize}

\end{section}

\begin{section}{Related Work}

\begin{paragraph}{Deep Object Detectors.}
Deep learning based object detection methods have shown dramatic improvements in both accuracy and speed recently. As one of the predominant detectors, two-stage detection methods~\cite{rcnn,fast-rcnn,faster-rcnn,rfcn,cai2018cascade,li2017light} first generate a set of region proposals and then refine them by CNN networks. In~\cite{rcnn}, R-CNN generates region proposals by Selective Search~\cite{uijlings2013selective} and then classifies and refines the cropped proposal regions from the original image by a standard CNN independently and sequentially. To reduce the redundant computation of feature extraction in R-CNN, SPPNet~\cite{he2015spatial} and Fast R-CNN~\cite{fast-rcnn} extract the feature of the whole image once, and then generate region features through spatial pyramid pooling and RoIPooling layers, respectively. The RoIPooling layer is further improved by RoIAlign layer~\cite{he2017mask} to address the coarse spatial quantization problem. Faster R-CNN~\cite{faster-rcnn} first proposes a unified end-to-end framework for object detection. It introduces a region proposal network (RPN) which shares the same backbone network with the detection network to replace the original standalone time-consuming region proposal methods. 
To further improve the efficiency of Faster R-CNN, R-FCN~\cite{rfcn} constructs position-sensitive score maps through a fully convolutional network to avoid the RoI-wise head network. To avoid additional large score maps in R-FCN, Light-Head R-CNN~\cite{li2017light} uses a thin feature map and a cheap R-CNN subnet to build a two-stage detector more efficiently.

On the other hand, one-stage methods which are popularized by YOLO~\cite{yolo,redmon2017yolo9000,redmon2018yolov3} and SSD~\cite{ssd}, aim to be more efficient by directly classifying the pre-defined anchors and further refining them using CNNs without the proposal generation step. Based on the multi-layer prediction module in SSD, DSSD~\cite{fu2017dssd} introduces additional context information with deconvolutional operators to improve the accuracy. RetinaNet~\cite{lin2017focal} proposes a new focal loss to address the extreme foreground-background class imbalance which stands out as a central issue in one-stage detectors. Inheriting the merits of two-stage approaches, RefineDet~\cite{zhang2018single} proposes an anchor refinement module to first filter the negative anchor boxes and coarsely adjust the anchor boxes for the next detection module.
\end{paragraph}

\begin{paragraph}{Methods for handling scale variation.}
As the most challenging problem in object detection, large scale variation across object instances hampers the accuracy of detectors. The multi-scale image pyramid~\cite{huang2017speed,liu2018path,deformable} is a common scheme to improve the detection methods, especially for objects of small and large scales. Based on the image pyramid strategy, SNIP~\cite{snip} proposes a scale normalization method to train objects that fall into the desired scale range for each resolution during multi-scale training. To perform multi-scale training more efficiently, SNIPER~\cite{sniper} only selects context regions around the ground-truth instances and sampled background regions for each scale during training. However, SNIP and SNIPER still suffer from the inevitable increase of inference time. 

Instead of taking multiple images as input, some methods utilize multi-level features of different spatial resolutions to alleviate scale variation. Methods like HyperNet~\cite{kong2016hypernet} and ION~\cite{bell2016inside} concatenate low-level and high-level features from different layers to generate better feature maps for prediction. Since the features from different layers usually have different resolutions, specific normalization or transformation operators need to be designed before fusing multi-level features. Instead, SSD~\cite{ssd} and MS-CNN~\cite{cai2016unified} perform object detection at multiple layers for objects of different scales without feature fusion.  TDM~\cite{shrivastava2016beyond} and FPN~\cite{fpn} further introduce a top-down pathway and lateral connections to enhance the semantic representation of low-level features at bottom layers. PANet~\cite{liu2018path} enhances the feature hierarchies in FPN by additional bottom-up path augmentation and proposes adaptive feature pooling to aggregate features from all  levels for better prediction. Rather than using features from different layers, our proposed TridentNet generates scale-specific features through multiple parallel branches, thus endowing our network the same representational power for objects of different scales.
\end{paragraph}

\begin{paragraph}{Dilated convolution.} Dilated convolution~\cite{yu2016multi} (aka Atrous convolution~\cite{holschneider1990real}) enlarges the convolution kernel with original weights by performing convolution at sparsely sampled locations, thus increasing the receptive field size without additional cost. Dilated convolution has been widely used in semantic segmentation to incorporate large-scale context information~\cite{yu2016multi,chen2018deeplab,zhao2017pyramid,chen2017rethinking}. In the object detection field, DetNet~\cite{li2018detnet} designs a specific detection backbone network to maintain the spatial resolution and enlarge the receptive field using dilated convolution. Deformable convolution~\cite{deformable} further generalizes dilated convolution by learning the sampling location adaptively. In our work, we employ dilated convolution in our multi-branch architecture with different dilation rates to adapt the receptive fields for objects of different scales.
\end{paragraph}

\end{section}

\begin{section}{Investigation of Receptive Field}\label{sec:pilot}
There are several design factors of the backbone network that may affect the performance of object detectors including downsample rate, network depth, and the receptive field. 
Several works~\cite{cai2016unified,li2018detnet} have already discussed their impacts. 
The effects of the first two factors are straightforward: deeper network with low downsample rate may increase the complexity, but benefit the detection task in general. 
Nevertheless, as far as we know, there is no previous work that studies the impact of the receptive field in isolation.

To investigate the effect of the receptive field on the detection of objects with different scales, we replace some convolutions in the backbone network with their dilated variants~\cite{yu2016multi}. 
We use different dilation rates to control the receptive field of a network.

Dilated convolution with a dilation rate $d_s$ inserts $d_s - 1$ zeros between consecutive filter values, enlarging the kernel size without increasing the number of parameters and computations. 
Specifically, a dilated 3$\times$3 convolution could have the same receptive field as the convolution with the kernel size of $3 + 2(d_s - 1)$. 
Suppose the total stride of current feature map is $s$ , then a dilated convolution of rate $d_s$ could increase the receptive field of the network by $2(d_s - 1)s$. 
Thus if we modify $n$ conv layers with $d_s$ dilation rate, the receptive field could be increased by $2(d_s - 1)sn$.

We conduct our pilot experiment using a Faster R-CNN~\cite{faster-rcnn} detector with the ResNet-C4 backbone on the COCO~\cite{coco} dataset. 
The results are reported in the COCO-style mmAP on all objects and objects of small, medium and large sizes, respectively. 
We use ResNet-50 and ResNet-101 as the backbone networks and vary the dilation rate $d_s$ of the 3$\times$3 convolutions from 1 to 3 for the residual blocks in the \textit{conv}$_4$ stage. 

Table~\ref{tb:dilation_exp} summarizes the results. We can find that as the receptive field increases (larger dilation rate), the performance of the detector on small objects drops consistently on both ResNet-50 and ResNet-101. While for large objects, the detector benefits from the increasing receptive fields. The above findings suggest that:

\begin{table}[hbpt]
\small
\renewcommand\arraystretch{1.1}
	\begin{center}
	\begin{tabular}{lc|c|ccc}
		\hline
		Backbone	& Dilation			& AP  & $\text{AP}_{s}$ & $\text{AP}_{m}$ & $\text{AP}_{l}$	\\
		\hline
		\multirow{3}{*}{ResNet-50}		& 1 & 0.332	& \textbf{0.174} & 0.384 & 0.464 \\
									& 2 & {0.342}	& 0.168	& \textbf{0.386} & 0.486 \\
									& 3 & 0.341	& 0.162	& 0.383	& \textbf{0.492} \\
		\hline
		\multirow{3}{*}{ResNet-101} 	& 1	& 0.379	& \textbf{0.200}	& \textbf{0.430}	& 0.528 \\
									& 2 & 0.380	& 0.191 & 0.427	& \textbf{0.538} \\
									& 3 & 0.371 & 0.181 & 0.410	& \textbf{0.538} \\		
		\hline						
	\end{tabular}
	\end{center}
\vspace{-10pt}
\caption{Object detection results with different receptive fields using Faster R-CNN~\cite{faster-rcnn} evaluated on the COCO \textit{minival} dataset~\cite{coco}.}\label{tb:dilation_exp}
\end{table}

\begin{enumerate}
\item The performance on objects of  different scales are influenced by the receptive field of a network. The most suitable receptive field is strongly correlated with the scale of objects.
\item Although ResNet-101 has a large enough theoretical receptive field to cover large objects (greater than 96$\times$96 resolution) in COCO, the performance of large objects could still be improved when enlarging the dilation rate. This finding shares the same spirit in~\cite{luo2016understanding} that the effective receptive field is smaller than the theoretical receptive field. We hypothesize that the effective receptive field of detection networks needs to balance between small and large objects. Increasing dilation rates enlarges the effective receptive field by emphasizing large objects, thus compromising the performance of small objects.
\end{enumerate}

The aforementioned experiments motivate us to adapt the receptive field for  objects of different scales as detailed in the next section.

\end{section}

\begin{section}{Trident Network}
\begin{figure*}[t]
\center{
\includegraphics[width=0.85\linewidth]{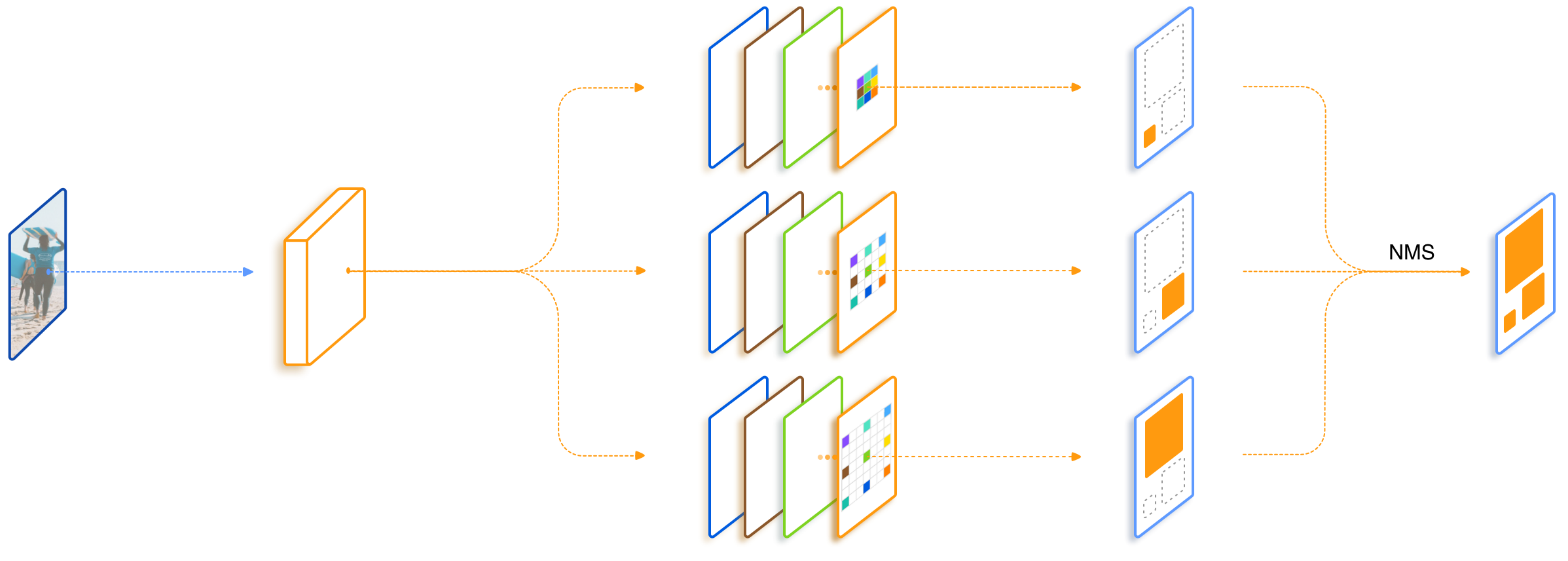}
\caption{Illustration of the proposed TridentNet. The multiple branches in trident blocks share the same parameters with different dilation rates to generate scale-specific feature maps. Objects of specified scales are sampled for each branch during training. The final proposals or detections from multiple branches will be combined using Non-maximum Suppression (NMS). Here we only show the backbone network of TridentNet. The RPN and Fast R-CNN heads are shared among branches and ignored for simplicity. }\label{fig:flowchart}}
\end{figure*}
In this section, we describe our scale-aware Trident Network (TridentNet) for object detection. The proposed TridentNet consists of weight sharing trident blocks and a deliberately designed scale-aware training scheme. Finally, We also present the inference details of TridentNet, including a fast inference approximation method.

\begin{subsection}{Network Structure}
Our goal is to inherit the merits of different receptive field sizes and avoid their drawbacks for detection networks. 
We propose a novel Trident architecture for this goal as shown in Figure~\ref{fig:flowchart}.
 In particular, our method takes a single-scale image as input, and then creates scale-specific feature maps through parallel branches where convolutions share the same parameters but with different dilation rates. 

\begin{paragraph}{Multi-branch Block}
We construct TridentNets by replacing some convolution blocks with the proposed \emph{trident blocks} in the backbone network of a detector.  
A trident block consists of multiple parallel branches in which each shares the same structure with the original convolution block except the dilation rate. 

Taking ResNet as an example, for a single residual block in the bottleneck style~\cite{resnet}, which consists of three convolutions with kernel size 1$\times$1, 3$\times$3 and 1$\times$1, a corresponding trident block is constructed as multiple parallel residual blocks with different dilation rates for the 3$\times$3 convs, as shown in Figure~\ref{fig:tridentblock}. Stacking trident blocks allows us to control receptive fields of different branches in an efficient way similar to the pilot experiment in Section~\ref{sec:pilot}.  
Typically, we replace the blocks in the last stage of the backbone network with trident blocks since larger strides lead to a larger difference in receptive fields as needed. 
Detailed design choices could be referred in Section~\ref{sec:exp_ablation}.
\end{paragraph}

\begin{paragraph}{Weight sharing among branches.}
An immediate problem of our multi-branch trident block is that it introduces several times parameters which may potentially incur overfitting. 
Fortunately, different branches share the same structure (except dilation rates) and thus make weight sharing straightforward.
In this work, we share the weights of all branches and their associated RPN and R-CNN heads, and only vary the dilation rate of each branch.

The advantages of weight sharing are three-fold. 
It reduces the number of parameters and makes TridentNet need no extra parameters compared with the original detector. 
It also echoes with our motivation that objects of different scales should go through a uniform transformation with the same representational power. 
A final point is that transformation parameters could be trained on more object samples from all branches. In other words, the same parameters are trained for different scale ranges under different receptive fields. 
\end{paragraph}
\end{subsection}

\begin{figure}[t]
	\centering
	\includegraphics[width=0.48\textwidth]{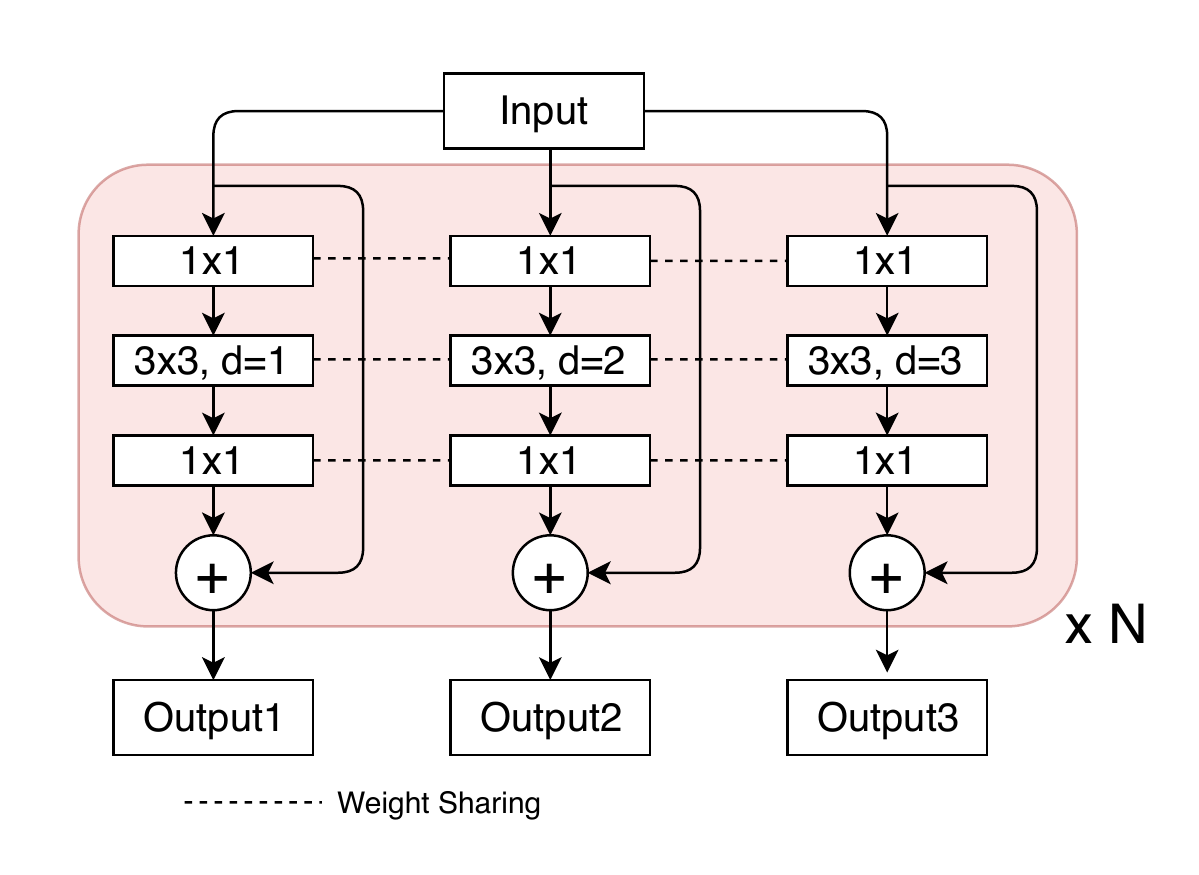}
	\caption{A trident block constructed from a bottleneck residual block.}
	\label{fig:tridentblock}
\end{figure}

\begin{subsection}{Scale-aware Training Scheme}
The proposed TridentNet architecture generates scale-specific feature maps according to the pre-defined dilation rates. 
However, the degradation in Table~\ref{tb:dilation_exp} caused by scale mismatching (\eg small objects on the branch with too large dilation) still exists for every single branch. 
Thus, it is natural to detect objects of different scales on different branches. 
Here, we propose a scale-aware training scheme to improve the scale awareness of every branch and avoid training objects of extreme scales on mismatched branches. 

Similar to SNIP~\cite{snip}, we define a valid range $[l_i, u_i]$ for each branch $i$. During training, we only select the proposals and ground truth boxes whose scales fall in the corresponding valid range of each branch. Specifically, for an Region-of-Interest (RoI) with width $w$ and height $h$ on the input image(before resize), it is valid for branch $i$ when:
\begin{equation}\label{eq:valid}
\begin{aligned}
l_i \leq \sqrt{wh} \leq u_i.
\end{aligned}
\end{equation}
This scale-aware training scheme could be applied on both RPN and R-CNN. 
For RPN, we select ground truth boxes which are valid for each branch according to Eq.~\ref{eq:valid} during anchor label assignment. 
Similarly, we remove all invalid proposals for each branch during the training of R-CNN. 
\end{subsection}

\begin{subsection}{Inference and Approximation}
During inference, we generate detection results for all branches and then filter out the boxes which fall outside the valid range of each branch. 
We  then use NMS or soft-NMS~\cite{softnms} to combine the detection outputs of multiple branches and obtain the final results. 

\begin{paragraph}{Fast Inference Approximation}
A major drawback of TridentNet is the slow inference speed due to its branching nature. 
Here we propose \emph{TridentNet Fast}, a fast approximation of TridentNet with only one branch during inference. 
For a three-branch network as in Figure~\ref{fig:flowchart}, we use the middle branch for inference since its valid range covers both large and small objects. 
In this way, TridentNet Fast incurs no additional time cost compared with a standard Faster R-CNN detector. 
Surprisingly, we find that this approximation only exhibits a slight performance drop compared with the original TridentNet. This may due to our weight-sharing strategy, through which multi-branch training is equivalent to within-network scale augmentation.
Detailed ablation of TridentNet Fast could be found in Section~\ref{sec:fast_infer}. 
\end{paragraph}
\end{subsection}

\end{section}

\begin{section}{Experiments}\label{sec:experiment}

In this section, we conduct experiments on the COCO dataset~\cite{coco}. 
Following~\cite{bell2016inside,fpn}, we train models on the union of 80k training images and 35k subset of validation images (\textit{trainval35k}), and evaluate on a set of 5k validation images (\textit{minival}). 
We also report the final results on a set of 20k test images (\textit{test-dev}). 
We first describe the implementation details of TridentNets and training settings in Section~\ref{sec:implementation}. 
We then conduct thorough ablation experiments to validate the proposed method in Section~\ref{sec:exp_ablation}. 
Finally, Section~\ref{sec:results} compares the results of TridentNets with state-of-the-art methods on the \textit{test-dev} set.

\begin{subsection}{Implementation Details}\label{sec:implementation}
We re-implement Faster R-CNN~\cite{faster-rcnn} as our baseline method in MXNet~\cite{mxnet}. 
Following other standard detectors~\cite{rcnn,faster-rcnn}, the network backbones are pre-trained on the ImageNet~\cite{imagenet}. The stem, the first residual stage, and all BN parameters are freezeed.  
The input images are resized to a short side of 800. Random horizontal flip is adopted during training.
By default, models are trained in a batch size of 16 on 8 GPUs. 
Models are trained in 12 epochs by default, with the learning rate starting from 0.02 and decreased by a factor of 0.1 after the 8th and 10th epoch. The $2\times$ or $3\times$ training schemes means doubling or tripling the total training epochs and learning rate schedules accordingly.
We adopt the output of \textit{conv}$_4$ stage in ResNet~\cite{resnet} as the backbone feature map and the \textit{conv}$_5$ stage as the R-CNN head in both baseline and TridentNet. 
If not otherwise specified, we adopt three branches as our default TridentNet structure.
For each branch in TridentNet, the top 12000/500 proposals are kept before/after NMS and we sample 128 ROIs for training. 
The dilation rates are set to 1, 2 and 3 in three branches, respectively. 
When adopting scale-aware training scheme for TridentNet, we set the valid ranges of three branches as $[0, 90]$, $[30, 160]$ and $[90, \infty]$, respectively.

For the evaluation, we report the standard COCO evaluation metric of Average Precision (AP)~\cite{coco} as well as $\text{AP}_{50}$/$\text{AP}_{75}$. 
We also report COCO-style $\text{AP}_s$, $\text{AP}_m$ and $\text{AP}_l$ on objects of small (less than 32$\times$32), medium (from 32$\times$32 to 96$\times$96) and large (greater than 96$\times$96) sizes. 
\end{subsection}

\begin{subsection}{Ablation Studies}\label{sec:exp_ablation}

\begin{paragraph}{Components of TridentNet.}

\begin{table*}[hbpt]
\small
\renewcommand\arraystretch{1.25}
	\begin{center}
	\setlength{\tabcolsep}{3.5pt}
	\begin{tabular}{l|l|ccc|cc|ccc}
		\hline
		Backbone & Method & Multi-branch & Weight-sharing & Scale-aware &  AP & $\text{AP}_{50}$ & $\text{AP}_{s}$ & $\text{AP}_{m}$ & $\text{AP}_{l}$	\\
		\hline
		\multirow{5}{*}{ResNet-101} & (a) Baseline & - & - & - 					&  	37.9	& 58.8	& {20.0}& {43.0}& 52.8\\
		\cline{2-10}
			& (b) Multi-branch 	 & \checkmark &	&								&	39.0	& 59.7	& 20.6	& 43.5	& 55.1\\
			& (c) TridentNet w/o scale-aware	& \checkmark &	\checkmark	&   &	40.3    & 61.1  & 21.8	& 44.7	& \textbf{56.7}\\
			& (d) TridentNet w/o sharing & \checkmark &	&	\checkmark 			&   39.3	& 60.4	& 21.4	& 43.8	& 54.2\\
			& (e) TridentNet	& \checkmark &	\checkmark & \checkmark			&	\textbf{40.6}	& \textbf{61.8}	& \textbf{23.0} & \textbf{45.5} & 55.9 \\
		\hline	
		\multirow{5}{*}{ResNet-101-Deformable} & (a) Baseline & - & - & - 		&  	39.9    & 61.3	& 21.6  & 45.0  & 55.6\\
		\cline{2-10}
			& (b) Multi-branch  & \checkmark &	&								&	40.5	& 61.5	& 21.9	& 45.3	& 56.8\\
			& (c) TridentNet w/o scale-aware	& \checkmark &	\checkmark 	&   &	41.4    & 62.8  & 23.4	& 45.9	& \textbf{57.4}\\
			& (d) TridentNet w/o sharing & \checkmark &	&	\checkmark 			&   40.3	& 61.6	& 22.9	& 45.0	& 55.0\\
			& (e) TridentNet	& \checkmark &	\checkmark & \checkmark			&	\textbf{41.8}	& \textbf{62.9}	& \textbf{23.6} & \textbf{46.8} & 57.1 \\
		\hline						
	\end{tabular}
	\end{center}
\vspace{-10pt}	
\caption{Results on the COCO \textit{minival} set. Starting from our baseline, we gradually add multi-branch design, weight sharing among branches and scale-aware training scheme in our TridentNet for ablation studies.}\label{tb:exp_ablation}
\end{table*}

First, we analyze the importance of each component in TridentNet. The baseline methods (Table~\ref{tb:exp_ablation}(a)) are evaluated on ResNet-101 and ResNet-101-Deformable~\cite{deformable} backbones. Then we gradually apply our multi-branch architecture, weight sharing design, and scale-aware training scheme. 
\begin{enumerate}
\item \textbf{Multi-branch}. Based on the pilot experiment, Table~\ref{tb:exp_ablation}(b) evaluates a straightforward way to get the best of multiple receptive fields. This multi-branch variant improves over the baselines on AP for both ResNet-101 (from 37.9 to 39.0) and ResNet-101-Deformable (from 39.9 to 40.5), especially for large objects (2.3/1.2 increase). This indicates that even the simplest multi-branch design could benefit from different receptive fields.
\item \textbf{Scale-aware}. Table~\ref{tb:exp_ablation}(d) shows the ablation results of adding scale-aware training based on multi-branch (Table~\ref{tb:exp_ablation}(b)). It brings additional improvements (0.8/1.0 increase on ResNet-101/ResNet-101-Deformable) for small objects but drops in performance for large objects. We conjecture that the scale-aware training design prevents each branch from training objects of extreme scales, but may also bring about the over-fitting problem in each branch caused by reduced effective samples.
\item \textbf{Weight-sharing}. By applying weight sharing on multi-branch (Table~\ref{tb:exp_ablation}(c)) and TridentNet (Table~\ref{tb:exp_ablation}(e)), we could achieve consistent improvements on both two base networks. This demonstrates the effectiveness of weight-sharing. It reduces the number of parameters and improves the performance of detectors. With the help of weight-sharing (Table~\ref{tb:exp_ablation}(e)), all branches share the same parameters which are fully trained on objects of all scales, thus alleviating the over-fitting issue in scale-aware training (Table~\ref{tb:exp_ablation}(d)). 
\end{enumerate}

Finally, TridentNets achieve significant improvements (2.7/1.9 AP increase) on the two base networks. It also reveals that the proposed TridentNet structure is compatible with methods like deformable convolution~\cite{deformable} which could adjust receptive field adaptively.

\end{paragraph}

\begin{paragraph}{Number of branches.}
We study the choice of the number of branches in TridentNets. Table~\ref{tb:nbranch_exp} shows the results using one to four branches. Note that we do not add scale-aware training scheme here to avoid elaborately tuning valid ranges for different numbers of branches. The results in Table~\ref{tb:nbranch_exp} demonstrate that TridentNets consistently improve over the single-branch method (baseline) with 2.7 to 3.4 AP increase. As can be noticed, four branches do not bring further improvement over three branches. Thus, considering the complexity and performance, we choose three branches as our default setting.

\begin{table}[t]
\small
\renewcommand\arraystretch{1.1}
	\begin{center}
	\begin{tabular}{c|cc|ccc}
		\hline
		Branches	& AP  & $\text{AP}_{50}$  & $\text{AP}_{s}$ & $\text{AP}_{m}$ & $\text{AP}_{l}$	\\
		\hline
		1 &  33.2	& 53.8	& 17.4	& 38.4	& 46.4 \\
		2 &  35.9	& 56.7	& \textbf{19.0}	& 40.6	& 51.2 \\
		3 &  \textbf{36.6}	& \textbf{57.3}	& 18.3	& \textbf{41.4}	& \textbf{52.3} \\
		4 &  36.5  & \textbf{57.3} & 18.8 & \textbf{41.4} & 51.9 \\
		\hline				
	\end{tabular}
	\end{center}
\vspace{-10pt}
\caption{Results on the COCO \textit{minival} set using different number of branches on ResNet-50.
}\label{tb:nbranch_exp}
\end{table}

\end{paragraph}

\begin{paragraph}{Stage of Trident blocks.}

We conduct ablation study on TridentNet to find the best stage to place trident blocks in ResNet. Table~\ref{tb:stage_exp} shows the results of applying trident blocks in \textit{conv}$_2$, \textit{conv}$_3$ and \textit{conv}$_4$ stages, respectively. The corresponding total strides are 4, 8 and 16. Comparing with \textit{conv}$_4$ stage, TridentNets on \textit{conv}$_2$ and \textit{conv}$_3$ stages achieve minor increase over the baseline. This is because the strides in \textit{conv}$_2$ and \textit{conv}$_3$ feature maps are not large enough to widen the discrepancy of receptive fields between three branches. 

\begin{table}[t]
\small
\renewcommand\arraystretch{1.1}
	\begin{center}
	\begin{tabular}{c|cc|ccc}
		\hline
		Stage	& AP  & $\text{AP}_{50}$  & $\text{AP}_{s}$ & $\text{AP}_{m}$ & $\text{AP}_{l}$	\\
		\hline
		Baseline &  33.2	& 53.8	& 17.4	& 38.4	& 46.4 \\
		\hline 
		conv2  &	  34.1	& 54.8	& 17.1	& 39.1	& 48.6  \\
		conv3  &  	  34.4  & 55.0  & 17.5	 & 39.3 & 49.0 \\
		conv4  &\textbf{36.6}	& \textbf{57.3}	& \textbf{18.3}	& \textbf{41.4}	& \textbf{52.3} \\
		\hline					
	\end{tabular}
	\end{center}
\vspace{-10pt}
\caption{Results on the COCO \textit{minival} set by replacing conv blocks with trident blocks in different stages of ResNet-50.
}\label{tb:stage_exp}
\end{table}

\end{paragraph}

\begin{paragraph}{Number of trident blocks.} 
As the \textit{conv}$_4$ stage in ResNet has multiple residual blocks, we also conduct ablation study to explore how many trident blocks are needed for TridentNet. Here we replace different numbers of residual blocks with trident blocks on \textit{conv}$_4$ on ResNet-101. The results in Figure~\ref{fig:number_blocks} show that when the number of trident blocks grows beyond 10, the performance of TridentNet becomes stable. This indicates the robustness of TridentNet with respect to the number of trident blocks, when the discrepancy of receptive fields between branches is large enough. 

\begin{figure}[hbpt]
\centering
\centering
\includegraphics[width=0.65\linewidth]{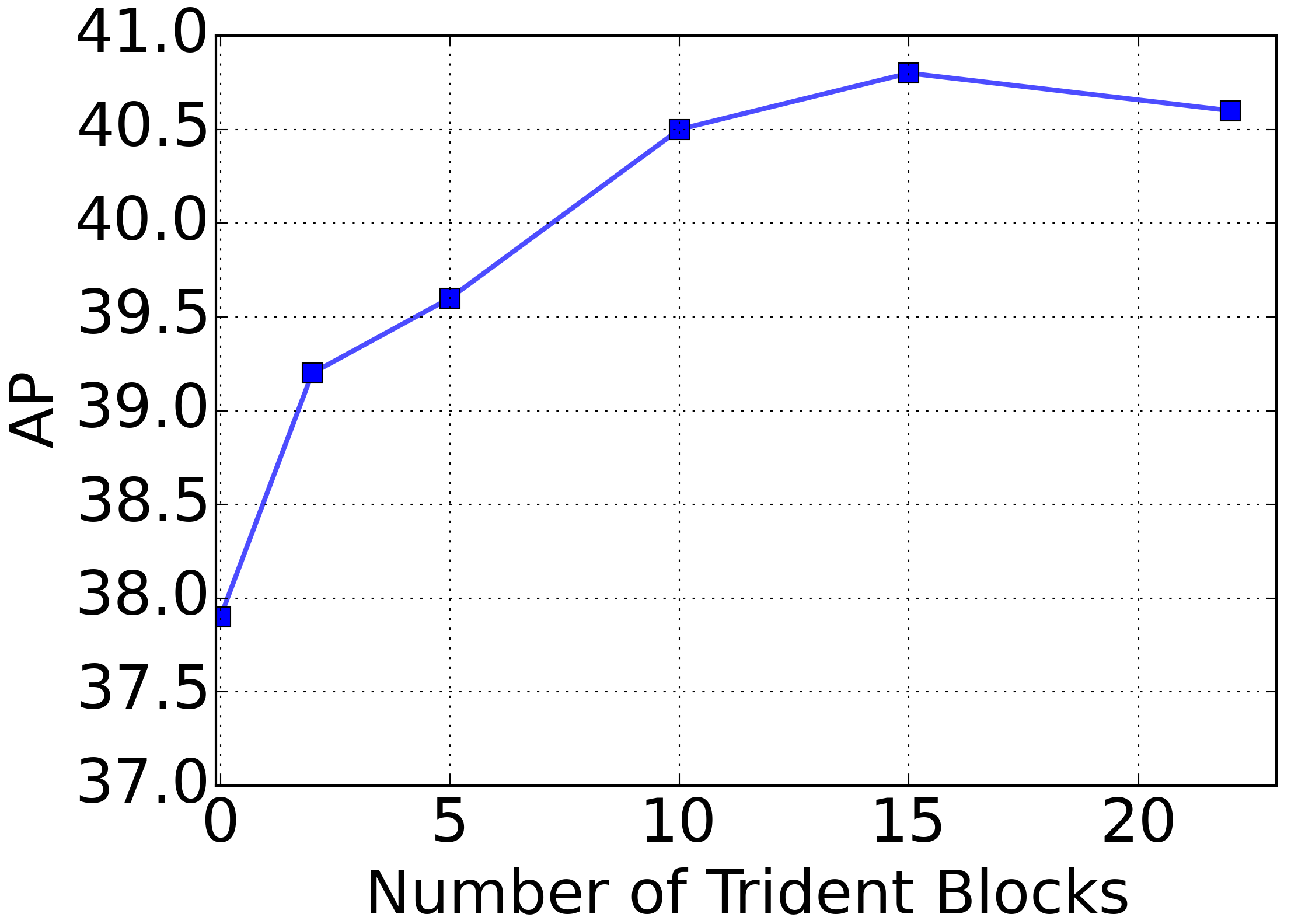}
\vspace{-1mm}
\caption{Results on the COCO \textit{minival} set using different number of trident blocks on ResNet-101.}\label{fig:number_blocks}
\end{figure}

\end{paragraph}

\begin{paragraph}{Performance of each branch.}
\label{sec:exp_single} 
In this section, we investigate the performance of each branch of our multi-branch TridentNet. 
We evaluate the performance of each branch independently without removing the detections out of the scale-aware range. 
Table~\ref{tb:branch_exp} shows the results of every single branch and three branches combined. 
As expected, through scale-aware training, branch-1 with the smallest receptive field achieves good results on small objects, branch-2 works well on objects of the medium scale while branch-3 with the largest receptive field is good at large objects. 
Finally, the three-branch method inherits the merits from three single branches and achieves the best results.
\end{paragraph}

\begin{table}[t]
\small
\renewcommand\arraystretch{1.1}
	\begin{center}
	\setlength{\tabcolsep}{3.5pt}
	\begin{tabular}{lc|cc|ccc}
		\hline
		Method & Branch No.	& AP  & $\text{AP}_{50}$  & $\text{AP}_{s}$ & $\text{AP}_{m}$ & $\text{AP}_{l}$	\\
		\hline
		Baseline & - & 	37.9	& 58.8	& {20.0}& {43.0}& 52.8\\
		\hline
		\multirow{4}{*}{TridentNet}		& Branch-1 & 31.5	& 53.9	& 22.0	& 43.3 & 29.9 \\
										& Branch-2 & 37.8	& 58.4	& 18.0	& 45.3 & 53.4 \\
										& Branch-3 & 31.9	& 48.8	& 7.1	& 37.9 & \textbf{56.1} \\
										& 3 Branches & \textbf{40.6}	& \textbf{61.8}	& \textbf{23.0} & \textbf{45.5} & 55.9 \\
		\hline
	\end{tabular}
	\end{center}
\vspace{-10pt}
\caption{Results of each branch in TridentNet evaluated on the COCO \textit{minival} set. The dilation rates of three branches in Trident Network are set as 1, 2 and 3. The results are based on ResNet-101.
}\label{tb:branch_exp}
\end{table}

\begin{table}[t]
\small
\renewcommand\arraystretch{1.1}
	\begin{center}
	\setlength{\tabcolsep}{3.5pt}
	\begin{tabular}{l|cc|ccc}
		\hline
		Scale-aware Ranges & AP  & $\text{AP}_{50}$  & $\text{AP}_{s}$ & $\text{AP}_{m}$ & $\text{AP}_{l}$	\\
		\hline
		(a) Baseline  & 	37.9	& 58.8	& {20.0}& {43.0}& 52.8\\
		\hline
		(b) $[0, 90], [30, 160], [90, \infty]$	 		& 37.8	& 58.4	& 18.0	& \textbf{45.3}  & 53.4 \\
		(c) $[0, 90], \enspace [0,  \infty], \enspace [90, \infty]$	 	& 39.3	& 60.1	& 19.1	& 44.6 & 56.4 \\
		(d) $[0, \infty], \enspace [0, \infty], \enspace [0, \infty]$  	& \textbf{40.0}	& \textbf{61.1}	& \textbf{20.9}	& 44.3	& \textbf{56.6} \\
		\hline
	\end{tabular}
	\end{center}
\vspace{-10pt}
\caption{Results of TridentNet Fast under different scale-aware range schemes evaluated on the COCO \textit{minival} set. All results are based on ResNet-101 and share the same hyper-parameters.}\label{tb:range_exp}
\end{table}
\end{subsection}

\begin{subsection}{Fast Inference Approximation}\label{sec:fast_infer}
To reduce the inference time of TridentNet, we propose TridentNet Fast which uses a single major branch to approximate the three-branch results during inference. As indicated in Table~\ref{tb:branch_exp}, branch-2 emerges as a natural candidate for the major branch as its scale-aware range covers most objects. We investigate the effect of scale-aware ranges for scale-aware training in Table~\ref{tb:range_exp}. As shown in Table~\ref{tb:range_exp}(c), by enlarging the scale-aware range of the major branch to incorporate objects of all scales, the performance of Trident Fast improves by 1.5 AP over the default scale-aware range setting. Furthermore, extending scale-aware ranges for all three branches achieves the best performance of 40.0 AP, which is close to the original TridentNet result of 40.6 AP. We hypothesize this may due to the weight-sharing strategy. Since the weights of the major branch are shared on other branches, training all branches in the scale-agnostic scheme is equivalent to perform within-network multi-scale augmentation. 

\end{subsection}

\begin{subsection}{Comparison with State-of-the-Arts}\label{sec:results}

\begin{table*}[t]
\small
\renewcommand\arraystretch{1.1}
	\begin{center}
	\begin{tabular}{ll|ccc|ccc}
		\hline
		Method &	Backbone	& AP  & $\text{AP}_{50}$  & $\text{AP}_{75}$ & $\text{AP}_{s}$ & $\text{AP}_{m}$ & $\text{AP}_{l}$	\\
		\hline
		Cascade R-CNN~\cite{cai2018cascade} 	& ResNet-101-FPN 			& 42.8 & 62.1 & 46.3 & 23.7	& 45.5 & 55.2 \\
		DCNv2~\cite{zhu2018deformable} & ResNet-101-DeformableV2                             & 46.0 & 67.9 & 50.8 & 27.8 & 49.1 & 59.5 \\
		DCR~\cite{cheng2018revisiting}			& ResNet-101-FPN-Deformable	& 43.1 & 66.1 & 47.3 & 25.8 & 45.9 & 55.3 \\
		SNIP~\cite{snip}	& ResNet-101-Deformable 						& 44.4 & 66.2 & 44.9 & 27.3 & 47.4 & 56.9 \\
		SNIPER~\cite{sniper}	& ResNet-101-Deformable 					& 46.1 & 67.0 & 51.6 & 29.6 & 48.9 & 58.1 \\
		\hline
		TridentNet						& ResNet-101						& 42.7 & 63.6 & 46.5 & 23.9 & 46.6 & 56.6\\
		TridentNet* 					& ResNet-101-Deformable 			& 46.8 & 67.6 & 51.5 & 28.0 & 51.2 & \textbf{60.5}\\ 
		TridentNet* + Image Pyramid	    & ResNet-101-Deformable	            & \textbf{48.4} & \textbf{69.7} & \textbf{53.5} & \textbf{31.8} & \textbf{51.3} & 60.3\\ 
		\hline			
	\end{tabular}
	\end{center}
\vspace{-10pt}
\caption{Comparisons of single-model results for different object detection methods evaluated on the COCO \textit{test-dev} set.
}\label{tb:results}
\end{table*}

In this section, we evaluate TridentNet on COCO \textit{test-dev} set and compare with other state-of-the-art methods. 
Here we report the results of our methods under different settings in Table~\ref{tb:results}.

TridentNet, which is to directly apply our method on Faster R-CNN with ResNet-101 backbone in the 2$\times$ training scheme, achieves 42.7 AP without bells and whistles. 

To fairly compare with SNIP and SNIPER, we apply multi-scale training, soft-NMS, deformable convolutions, large-batch BN, and the 3$\times$ training scheme on TridentNet and get TridentNet*. 
It gives an AP of 46.8, already surpassing image pyramid based SNIP and SNIPER in the single-scale testing setting. 
If we adopt the image pyramid for testing, it further improves the result of TridentNet* to 48.4 AP. 
To our best knowledge, for single models with ResNet-101 backbone, our result is the best entry among state-of-the-art methods. 
Furthermore, TridentNet* Fast + Image Pyramid achieves 47.6 AP.

\begin{paragraph}{Compare with other scale handling methods.} 
In this section, we compare TridentNet with other popular scale handling methods like FPN~\cite{fpn} and ASPP~\cite{chen2017rethinking}. 
FPN is the \emph{de facto} model for handling scale variation in detection.
ASPP is a special case of TridentNet with only one trident block and dilation rates of three branches set to (6, 12, 18), followed by a feature fusion operator. 
To fairly compare with FPN, we adopt a \emph{2fc} head instead of a \emph{conv5} head for models in this section. 
Table~\ref{tb:fpn_compare} compares these methods under the same training setting. 
TridentNet improves significantly over other methods on all scales. 
It shows the effectiveness of scale specific feature maps generated by TridentNet with the same set of parameters. 
Furthermore, TridentNet Fast achieves 41.0 AP which improves 1.2 AP over the baseline with no computation cost.
{\let\thefootnote\relax\footnote{$^\dagger$Detectron: \url{https://github.com/facebookresearch/Detectron/blob/master/MODEL_ZOO.md}}}

\begin{table}[t]
\small
\renewcommand\arraystretch{1.1}
	\begin{center}
	\setlength{\tabcolsep}{3.5pt}
	\begin{tabular}{l|ccc|ccc}
		\hline
		Method &  AP  & $\text{AP}_{50}$  &  $\text{AP}_{75}$ & $\text{AP}_{s}$ & $\text{AP}_{m}$ & $\text{AP}_{l}$	\\
		\hline
		\emph{2fc} Baseline & 39.8 & 61.7 & 43.0 & 22.0 & 44.7 & 54.4 \\
		FPN$^\dagger$~\cite{fpn} & 39.8 & 61.3 & 43.3 & 22.9 & 43.3 & 52.6 \\
		ASPP & 39.7 & 60.4 & 42.7 & 21.7 & 44.5 & 53.9 \\
		TridentNet & \textbf{42.0} & \textbf{63.5} & \textbf{45.5} & \textbf{24.9} & \textbf{47.0} & \textbf{56.9} \\
		\hline
	\end{tabular}
	\end{center}
\vspace{-10pt}
\caption{Comparisons of detection results on the COCO \textit{minival} set. Following FPN$^\dagger$, all methods are based on ResNet-101 with \emph{2fc} head using 2$\times$ training schedule.
}\label{tb:fpn_compare}
\end{table}

\end{paragraph}

\end{subsection}

\end{section}

\begin{section}{Conclusion}
In this paper, we present a simple object detection method called Trident Network to build in-network scale-specific feature maps with the uniform representational power. A scale-aware training scheme is adopted for our multi-branch architecture to equip each branch with the specialized ability for corresponding scales. The fast inference method with the major branch makes TridentNet achieve significant improvements over baseline methods without any extra parameters and computations. 
\end{section}

\section*{Acknowledge}
\vspace{-5pt}
This work was supported in part by the National Key R\&D Program of China (No.2018YFB1402605), the Beijing Municipal Natural Science Foundation (No.Z181100008918010), the National Natural Science Foundation of China (No.61836014, No.61761146004, No.61773375, No.61602481). The authors would like to thanks NVAIL for the support.

{\small
\bibliographystyle{ieee_fullname}
\bibliography{egbib}

\begin{thebibliography}{10}\itemsep=-1pt

\bibitem{adelson1984pyramid}
Edward~H Adelson, Charles~H Anderson, James~R Bergen, Peter~J Burt, and Joan~M
  Ogden.
\newblock Pyramid methods in image processing.
\newblock {\em RCA engineer}, 29(6):33--41, 1984.

\bibitem{bell2016inside}
Sean Bell, C Lawrence~Zitnick, Kavita Bala, and Ross Girshick.
\newblock Inside-outside net: Detecting objects in context with skip pooling
  and recurrent neural networks.
\newblock In {\em CVPR}, 2016.

\bibitem{softnms}
Navaneeth Bodla, Bharat Singh, Rama Chellappa, and Larry~S Davis.
\newblock {Soft-NMS}-improving object detection with one line of code.
\newblock In {\em ICCV}, 2017.

\bibitem{cai2016unified}
Zhaowei Cai, Quanfu Fan, Rogerio~S Feris, and Nuno Vasconcelos.
\newblock A unified multi-scale deep convolutional neural network for fast
  object detection.
\newblock In {\em ECCV}, 2016.

\bibitem{cai2018cascade}
Zhaowei Cai and Nuno Vasconcelos.
\newblock Cascade {R-CNN}: Delving into high quality object detection.
\newblock In {\em CVPR}, 2018.

\bibitem{chen2018deeplab}
Liang-Chieh Chen, George Papandreou, Iasonas Kokkinos, Kevin Murphy, and Alan~L
  Yuille.
\newblock Deeplab: Semantic image segmentation with deep convolutional nets,
  atrous convolution, and fully connected crfs.
\newblock {\em IEEE Transactions on Pattern Analysis and Machine Intelligence},
  40(4):834--848, 2018.

\bibitem{chen2017rethinking}
Liang-Chieh Chen, George Papandreou, Florian Schroff, and Hartwig Adam.
\newblock Rethinking atrous convolution for semantic image segmentation.
\newblock {\em arXiv:1706.05587}, 2017.

\bibitem{mxnet}
Tianqi Chen, Mu Li, Yutian Li, Min Lin, Naiyan Wang, Minjie Wang, Tianjun Xiao,
  Bing Xu, Chiyuan Zhang, and Zheng Zhang.
\newblock {MXNet}: A flexible and efficient machine learning library for
  heterogeneous distributed systems.
\newblock In {\em NIPS Workshop}, 2015.

\bibitem{cheng2018revisiting}
Bowen Cheng, Yunchao Wei, Honghui Shi, Rogerio Feris, Jinjun Xiong, and Thomas
  Huang.
\newblock Revisiting rcnn: On awakening the classification power of faster
  rcnn.
\newblock In {\em ECCV}, 2018.

\bibitem{rfcn}
Jifeng Dai, Yi Li, Kaiming He, and Jian Sun.
\newblock {R-FCN}: Object detection via region-based fully convolutional
  networks.
\newblock In {\em NIPS}, 2016.

\bibitem{deformable}
Jifeng Dai, Haozhi Qi, Yuwen Xiong, Yi Li, Guodong Zhang, Han Hu, and Yichen
  Wei.
\newblock Deformable convolutional networks.
\newblock In {\em ICCV}, 2017.

\bibitem{dalal2005histograms}
Navneet Dalal and Bill Triggs.
\newblock Histograms of oriented gradients for human detection.
\newblock In {\em CVPR}, 2005.

\bibitem{dollar2014fast}
Piotr Doll{\'a}r, Ron Appel, Serge Belongie, and Pietro Perona.
\newblock Fast feature pyramids for object detection.
\newblock {\em IEEE Transactions on Pattern Analysis and Machine Intelligence},
  36(8):1532--1545, 2014.

\bibitem{fu2017dssd}
Cheng-Yang Fu, Wei Liu, Ananth Ranga, Ambrish Tyagi, and Alexander~C Berg.
\newblock Dssd: Deconvolutional single shot detector.
\newblock {\em arXiv:1701.06659}, 2017.

\bibitem{fast-rcnn}
Ross Girshick.
\newblock Fast {R-CNN}.
\newblock In {\em ICCV}, 2015.

\bibitem{rcnn}
Ross Girshick, Jeff Donahue, Trevor Darrell, and Jitendra Malik.
\newblock Rich feature hierarchies for accurate object detection and semantic
  segmentation.
\newblock In {\em CVPR}, 2014.

\bibitem{he2017mask}
Kaiming He, Georgia Gkioxari, Piotr Doll{\'a}r, and Ross Girshick.
\newblock Mask {R-CNN}.
\newblock In {\em ICCV}, 2017.

\bibitem{he2015spatial}
Kaiming He, Xiangyu Zhang, Shaoqing Ren, and Jian Sun.
\newblock Spatial pyramid pooling in deep convolutional networks for visual
  recognition.
\newblock {\em IEEE Transactions on Pattern Analysis and Machine Intelligence},
  37(9):1904--1916, 2015.

\bibitem{resnet}
Kaiming He, Xiangyu Zhang, Shaoqing Ren, and Jian Sun.
\newblock Deep residual learning for image recognition.
\newblock In {\em CVPR}, 2016.

\bibitem{holschneider1990real}
Matthias Holschneider, Richard Kronland-Martinet, Jean Morlet, and Ph
  Tchamitchian.
\newblock A real-time algorithm for signal analysis with the help of the
  wavelet transform.
\newblock In {\em Wavelets}, pages 286--297. 1990.

\bibitem{huang2017speed}
Jonathan Huang, Vivek Rathod, Chen Sun, Menglong Zhu, Anoop Korattikara,
  Alireza Fathi, Ian Fischer, Zbigniew Wojna, Yang Song, Sergio Guadarrama,
  et~al.
\newblock Speed/accuracy trade-offs for modern convolutional object detectors.
\newblock In {\em CVPR}, 2017.

\bibitem{kong2016hypernet}
Tao Kong, Anbang Yao, Yurong Chen, and Fuchun Sun.
\newblock Hypernet: Towards accurate region proposal generation and joint
  object detection.
\newblock In {\em CVPR}, 2016.

\bibitem{li2017light}
Zeming Li, Chao Peng, Gang Yu, Xiangyu Zhang, Yangdong Deng, and Jian Sun.
\newblock Light-head {R-CNN}: In defense of two-stage object detector.
\newblock {\em arXiv:1711.07264}, 2017.

\bibitem{li2018detnet}
Zeming Li, Chao Peng, Gang Yu, Xiangyu Zhang, Yangdong Deng, and Jian Sun.
\newblock {DetNet}: Design backbone for object detection.
\newblock In {\em ECCV}, 2018.

\bibitem{fpn}
Tsung-Yi Lin, Piotr Doll{\'a}r, Ross~B Girshick, Kaiming He, Bharath Hariharan,
  and Serge~J Belongie.
\newblock Feature pyramid networks for object detection.
\newblock In {\em CVPR}, 2017.

\bibitem{lin2017focal}
Tsung-Yi Lin, Priya Goyal, Ross Girshick, Kaiming He, and Piotr Dollar.
\newblock Focal loss for dense object detection.
\newblock In {\em ICCV}, 2017.

\bibitem{coco}
Tsung-Yi Lin, Michael Maire, Serge Belongie, James Hays, Pietro Perona, Deva
  Ramanan, Piotr Doll{\'a}r, and C~Lawrence Zitnick.
\newblock Microsoft coco: Common objects in context.
\newblock In {\em ECCV}, 2014.

\bibitem{liu2018path}
Shu Liu, Lu Qi, Haifang Qin, Jianping Shi, and Jiaya Jia.
\newblock Path aggregation network for instance segmentation.
\newblock In {\em CVPR}, 2018.

\bibitem{ssd}
Wei Liu, Dragomir Anguelov, Dumitru Erhan, Christian Szegedy, Scott Reed,
  Cheng-Yang Fu, and Alexander~C Berg.
\newblock {SSD}: Single shot multibox detector.
\newblock In {\em ECCV}, 2016.

\bibitem{lowe2004distinctive}
David~G Lowe.
\newblock Distinctive image features from scale-invariant keypoints.
\newblock {\em International Journal of Computer Vision}, 60(2):91--110, 2004.

\bibitem{luo2016understanding}
Wenjie Luo, Yujia Li, Raquel Urtasun, and Richard Zemel.
\newblock Understanding the effective receptive field in deep convolutional
  neural networks.
\newblock In {\em NIPS}, 2016.

\bibitem{yolo}
Joseph Redmon, Santosh Divvala, Ross Girshick, and Ali Farhadi.
\newblock You only look once: Unified, real-time object detection.
\newblock In {\em CVPR}, 2016.

\bibitem{redmon2017yolo9000}
Joseph Redmon and Ali Farhadi.
\newblock {YOLO9000}: Better, faster, stronger.
\newblock In {\em CVPR}, 2017.

\bibitem{redmon2018yolov3}
Joseph Redmon and Ali Farhadi.
\newblock {YOLOv3}: An incremental improvement.
\newblock {\em arXiv:1804.02767}, 2018.

\bibitem{faster-rcnn}
Shaoqing Ren, Kaiming He, Ross Girshick, and Jian Sun.
\newblock Faster {R-CNN}: Towards real-time object detection with region
  proposal networks.
\newblock In {\em NIPS}, 2015.

\bibitem{imagenet}
Olga Russakovsky, Jia Deng, Hao Su, Jonathan Krause, Sanjeev Satheesh, Sean Ma,
  Zhiheng Huang, Andrej Karpathy, Aditya Khosla, Michael Bernstein,
  Alexander~C. Berg, and Li Fei-Fei.
\newblock {ImageNet} large scale visual recognition challenge.
\newblock {\em International Journal of Computer Vision}, 115(3):211--252,
  2015.

\bibitem{shrivastava2016beyond}
Abhinav Shrivastava, Rahul Sukthankar, Jitendra Malik, and Abhinav Gupta.
\newblock Beyond skip connections: Top-down modulation for object detection.
\newblock {\em arXiv:1612.06851}, 2016.

\bibitem{snip}
Bharat Singh and Larry~S Davis.
\newblock An analysis of scale invariance in object detection--{SNIP}.
\newblock In {\em CVPR}, 2018.

\bibitem{sniper}
Bharat Singh, Mahyar Najibi, and Larry~S Davis.
\newblock {SNIPER}: Efficient multi-scale training.
\newblock In {\em NIPS}, 2018.

\bibitem{uijlings2013selective}
Jasper~RR Uijlings, Koen~EA Van De~Sande, Theo Gevers, and Arnold~WM Smeulders.
\newblock Selective search for object recognition.
\newblock {\em International Journal of Computer Vision}, 104(2):154--171,
  2013.

\bibitem{yu2016multi}
Fisher Yu and Vladlen Koltun.
\newblock Multi-scale context aggregation by dilated convolutions.
\newblock In {\em ICLR}, 2016.

\bibitem{zhang2018single}
Shifeng Zhang, Longyin Wen, Xiao Bian, Zhen Lei, and Stan~Z Li.
\newblock Single-shot refinement neural network for object detection.
\newblock In {\em CVPR}, 2018.

\bibitem{zhao2017pyramid}
Hengshuang Zhao, Jianping Shi, Xiaojuan Qi, Xiaogang Wang, and Jiaya Jia.
\newblock Pyramid scene parsing network.
\newblock In {\em CVPR}, 2017.

\bibitem{zhu2018deformable}
Xizhou Zhu, Han Hu, Stephen Lin, and Jifeng Dai.
\newblock Deformable convnets v2: More deformable, better results.
\newblock {\em CVPR}, 2019.

\end{thebibliography}
}

\end{document}